\documentclass[a4,10pt]{article}
\usepackage{amsmath,amssymb, bm, statex, braket}
\usepackage{enumerate, fancybox}
\usepackage{mathtools}

\usepackage[dvipdfmx]{graphicx}

\allowdisplaybreaks[4]
\newcommand{\nn}{\nonumber \\}
\newcommand{\mcal}[1]{ \mathcal{#1} }
\newcommand{\mrm}[1]{ \mathrm{#1} }
\newcommand{\ve}[1]{ \mathbf{#1} }
\newcommand{\ave}[1]{\langle #1 \rangle}

\begin{document}

\begin{center}
{\Large{Restricted Boltzmann Machine with Multivalued Hidden Variables: a model suppressing over-fitting}}\\

\ \\
\ \\
{\large{
Yuuki Yokoyama$^*$, Tomu Katsumata$^{\dagger}$, and Muneki Yasuda$^{\dagger}$
}}\\
\ \\
$^*$ALBERT Inc., Japan\\
$^{\dagger}$Graduate School of Science and Engineering, Yamagata University, Japan
\end{center}

\subsubsection*{abstract:}
Generalization is one of the most important issues in machine learning problems. 
In this study, we consider generalization in restricted Boltzmann machines (RBMs). 
We propose an RBM with multivalued hidden variables,  
which is a simple extension of conventional RBMs. 
We demonstrate that the proposed model is better than the conventional model via numerical 
experiments for contrastive divergence learning with artificial data and a classification problem with MNIST.

\section{Introduction}
\label{sec:introduction}

Generalization is one of the most important goals in statistical machine learning problems~\cite{Bishop2006}. 
In various standard machine learning techniques, given a particular data set, 
we fit our probabilistic learning model to the empirical distribution (or the data distribution) of the data set. 
When our learning model is sufficiently flexible, it can fit the empirical distribution exactly via an appropriate learning method. 
A learning model that is too close to the empirical distribution frequently gives poor results for new data points. 
This situation is known as \textit{over-fitting}. 
Over-fitting impedes generalization; therefore, techniques that can suppress over-fitting are needed to achieve good generalizations.
Regularizations, such as $L_1$ and $L_2$ regularizations or their combination (the elastic net)~\cite{ElasticNet2005}, are popular techniques used for this purpose.  

Here, we focus on a restricted Boltzmann machine (RBM)~\cite{RBM1986,CD2002}. 
RBMs have a wide range of applications such as collaborating filtering~\cite{RBMcFilter2007}, 
classification~\cite{DRBM2008}, and deep learning~\cite{Hinton2006,DBM2009,DBM2012}.
The suppression of over-fitting is also important in RBMs.
An RBM is a probabilistic neural network defined on a bipartite undirected graph 
comprising two different layers: visible layer and hidden layer. 
The visible layer, which consists of visible (random) variables, directly corresponds to the data points,  
while the hidden layer, which consists of hidden (random) variables, does not. 
The hidden layer creates complex correlations among the visible variables. 
The sample space of the visible variables is determined by the range of data elements, 
whereas the sample space of the hidden variables can be set freely.
Typically, the hidden variables are given binary values ($\{0,1\}$ or $\{-1,+1\}$). 

In this study, we propose an RBM with multivalued hidden variables. 
The proposed RBM is a very simple extension of the conventional RBM with binary-hidden variables (referred to as the binary-RBM in this paper). 
However, we demonstrate that the proposed RBM is better than the binary-RBM in terms of suppressing the over-fitting.  
The remainder of this paper is organized as follows. 
We define the proposed RBM in Sec. \ref{sec:RBM} and explain its maximum likelihood estimation in Sec. \ref{sec:log-likelihood-function}. 
In Sec. \ref{sec:RBM_experiment}, we demonstrate the validity of the proposed RBM using numerical experiments for contrastive divergence (CD) learning~\cite{CD2002} with artificial data. 
We give an insight on the effect of our extension (i.e., the effect of multivalued hidden variables) 
using a toy example in Sec. \ref{sec:ToyExample}.
In Sec. \ref{sec:PatternRecognitionApplication}, we apply the proposed RBM to a classification problem 
and show that it is also effective in such type of problems. 
Finally, the conclusion is given in Sec. \ref{sec:conclusion}.

\section{Restricted Boltzmann machine with multivalued hidden variables}
\label{sec:RBM}

\begin{figure}[tb]
\centering
\includegraphics[height=1.5cm]{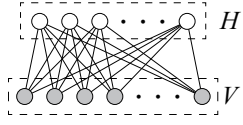}
\caption{Bipartite graph consisting of two layers: the visible layer and the hidden layer. 
$V$ and $H$ are the sets of indices of the nodes in the visible layer and hidden layer, respectively.}
\label{fig:RBM}
\end{figure}

Let us consider a bipartite graph consisting of two different layers: the visible layer and hidden layer, as shown in Fig. \ref{fig:RBM}. 
Binary (or bipolar) visible variables, $\bm{v} := \{v_i \in \{-1,+1\} \mid i \in V\}$, are assigned to the corresponding nodes in the visible layer. 
The corresponding hidden variables, $\bm{h} := \{h_j \in \mcal{X}(s) \mid j \in H\}$, are assigned to the nodes in the hidden layer, 
where $\mcal{X}(s)$ is the sample space defined by
\begin{align}
\mcal{X}(s):= \big\{ (2k - s)/s \mid k= 0,1,\ldots,s \big\} \quad s \in \mathbb{N},
\label{eqn:SampleSpace_of_hidden}
\end{align}
where $\mathbb{N}$ is the set of all natural numbers.
For example, $\mcal{X}(1) = \{-1,+1\}$, $\mcal{X}(2) = \{-1,0,+1\}$, and $\mcal{X}(3) = \{-1,-1/3, +1/3,+1\}$.
Namely, $\mcal{X}(s)$ is the set of values that evenly partition the interval $[-1,+1]$ into $(s+1)$ parts. 
We define that in the limit of $s \to \infty$, $\mcal{X}(s)$ becomes a continuous space $[-1, +1]$, i.e., $\mcal{X}(\infty) = [-1,+1]$. 
On the bipartite graph, we define the energy function for $s \in \mathbb{N}$ as
\begin{align}
E_s(\bm{v},\bm{h} ; \theta):=-\sum_{i \in V}b_i v_i-\sum_{j \in H}c_j h_j
-\sum_{i \in V}\sum_{j \in H}w_{i,j}v_i h_j,
\label{eqn:EnergyFunction}
\end{align}
where $\{b_i\}$, $\{c_j\}$, and $\{w_{i,j}\}$ are the learning parameters of the energy function, and they are collectively denoted by $\theta$. 
Specifically, $\{b_i\}$ and $\{c_j\}$ are the biases for the visible and hidden variables, respectively, and $\{w_{i,j}\}$ are the couplings between the visible and hidden variables.
Our RBM is defined in the form of a Boltzmann distribution in terms of the energy function given in Eq. (\ref{eqn:EnergyFunction}):  
\begin{align}
P_s(\bm{v}, \bm{h} \mid \theta):=\frac{\omega(s)}{Z_s(\theta)}  \exp\big(-E_s(\bm{v},\bm{h} ; \theta)\big),
\label{eqn:RBM}
\end{align}
where
\begin{align}
Z_s(\theta):=\sum_{\bm{v} \in \{-1,+1\}^{|V|}}\sum_{\bm{h} \in \mcal{X}(s)^{|H|}} \omega(s)\exp\big(-E_s(\bm{v},\bm{h} ; \theta)\big)
\label{eqn:PartitionFunction_RBM}
\end{align}
is the partition function. 
The multiple summations in Eq. (\ref{eqn:PartitionFunction_RBM}) mean 
\begin{align*}
\sum_{\bm{v} \in \{-1,+1\}^{|V|}} &= \sum_{v_1 \in \{-1,+1\}}\sum_{v_2 \in \{-1,+1\}}\cdots \sum_{v_{|V|} \in \{-1,+1\}}, \\
\sum_{\bm{h} \in \mcal{X}(s)^{|H|}} &= \sum_{h_1 \in \mcal{X}(s)}\sum_{h_2 \in \mcal{X}(s)}\cdots \sum_{h_{|H|} \in \mcal{X}(s)}.
\end{align*}
The factor $\omega(s):= \{2/(s + 1)\}^{|H|}$ appearing in Eqs. (\ref{eqn:RBM}) and (\ref{eqn:PartitionFunction_RBM}) is a constant unrelated to $\bm{v}$ and $\bm{h}$.   
Although it vanishes by reducing the fraction in Eq.~(\ref{eqn:RBM}), we leave it for the sake of the subsequent analysis. 
The factor lets the summation over $h_j$ be a Riemann sum and prevents the divergence of the partition function in $s \to \infty$.
It is noteworthy that when $s = 1$, Eq. (\ref{eqn:RBM}) is equivalent to the binary-RBM.

The marginal distribution of RBM is expressed as
\begin{align}
P_s(\bm{v}\mid \theta)&=\sum_{\bm{h} \in \mcal{X}(s)^{|H|}}P_s(\bm{v}, \bm{h} \mid \theta)\nn
&=\frac{1}{Z_s(\theta)}\exp\Big(\sum_{i \in V}b_i v_i + \sum_{j \in H}\ln \phi_s\big(\lambda_j(\bm{v},\theta)\big)\Big),
\label{eqn:MargialDistribution}
\end{align}
where $\lambda_j(\bm{v},\theta):= c_j + \sum_{i \in V}w_{i,j}v_i$ and $\phi_s(x):= \sum_{h \in \mcal{X}(s)}2(s+1)^{-1} e^{x h}$. 
It is noteworthy that factor $2(s + 1)^{-1}$ in the definition of $\phi_s(x)$ comes from $\omega(s)$.
Using the formula of geometric series, we obtain 
\begin{align}
\phi_s(x) =\frac{2\sinh \{(s+1)x/s\}}{(s + 1)\sinh (x/s)}
\label{eqn:phi_s(x)}
\end{align}
for $1 \leq s < \infty$. 
When $s \to \infty$, we obtain
\begin{align}
\phi_{\infty}(x) =\int_{-1}^{+1} e^{x h} dh = \frac{2 \sinh x}{x}.
\label{eqn:phi_inf(x)}
\end{align}
The additive factor, $2(s + 1)^{-1}$, ensures that $\lim_{s\to \infty}\phi_s(x) = \phi_{\infty}(x)$.
The conditional distributions are
\begin{align}
P_s(\bm{v} \mid \bm{h}, \theta)&=\prod_{i \in V}P_s(v_i \mid \bm{h}, \theta),\quad
P_s(v_i \mid \bm{h}, \theta) \propto \exp\big(\xi_i(\bm{h},\theta) v_i\big),
\label{eqn:CondDistribution_V|H}\\
P_s(\bm{h} \mid \bm{v}, \theta)&=\prod_{j \in H}P_s(h_j \mid \bm{v}, \theta),\quad
P_s(h_j \mid \bm{v}, \theta) \propto \exp\big(\lambda_j(\bm{v},\theta) h_j\big),
\label{eqn:CondDistribution_H|V}
\end{align}
where $\xi_i(\bm{h},\theta):= b_i + \sum_{j \in H}w_{i,j}h_j$. 
We can easily sample $\bm{v}$ from a given $\bm{h}$ using Eq. (\ref{eqn:CondDistribution_V|H}) 
and sample $\bm{h}$ from a given $\bm{v}$ using Eq. (\ref{eqn:CondDistribution_H|V}). 
Alternately repeating these two kinds of conditional samplings yields a (blocked) Gibbs sampling on the RBM. 
It is noteworthy that when $s \to \infty$, the conditional sampling of $\bm{h}$ using Eq. (\ref{eqn:CondDistribution_H|V}) 
can be implemented using the inverse transform sampling. 
The cumulative distribution function of $P_{\infty}(h_j \mid \bm{v}, \theta)$ is 
\begin{align*}
F(x):=\int_{-1}^x \frac{\exp\big(\lambda_j(\bm{v},\theta) h_j\big)}{\phi_{\infty}\big(\lambda_j(\bm{v},\theta)\big)} dh_j=\frac{\exp\big(\lambda_j(\bm{v},\theta) x\big)- \exp\big(-\lambda_j(\bm{v},\theta) \big)}{2 \sinh \lambda_j(\bm{v},\theta)},
\end{align*}
and therefore, its inverse function is 
\begin{align*}
F^{-1}(u):=\frac{1}{\lambda_j(\bm{v},\theta)} \ln \big\{\exp\big(-\lambda_j(\bm{v},\theta) \big) + 2 u \sinh \lambda_j(\bm{v},\theta)\big\}.
\end{align*}
$F^{-1}(u)$ is the sampled value of $h_j$ from $P_{\infty}(h_j \mid \bm{v}, \theta)$, where $u$ is a sample point from the uniform distribution over $[0,1]$.

\subsection{Log-likelihood function and its gradients}
\label{sec:log-likelihood-function}

Given $N$ training data points for the visible layer, $D_V:=\{ \ve{v}^{(\mu)} \in \{-1,+1\}^{|V|} \mid \mu = 1,2,\ldots,N\}$, 
the learning of RBM is done by maximizing the log-likelihood function (or the negative cross-entropy loss function), defined by
\begin{align}
l_s(\theta):=\frac{1}{N}\sum_{\mu = 1}^N \ln P_s(\ve{v}^{(\mu)} \mid \theta),
\label{eqn:LogLikelihood}
\end{align}
with respect to $\theta$ (namely, the maximum likelihood estimation).  
The distribution in the logarithmic function in Eq. (\ref{eqn:LogLikelihood}) is the marginal distribution obtained in Eq. (\ref{eqn:MargialDistribution}). 
The log-likelihood function is regarded as the negative training error.
Usually, the log-likelihood function is maximized using a gradient ascent method.
The gradients of the log-likelihood function with respect to the learning parameters are as follows.
\begin{align}
\frac{\partial l_s(\theta)}{\partial b_i}&=\frac{1}{N}\sum_{\mu = 1}^N \mrm{v}_i^{(\mu)} -\ave{v_i}_s,
\label{eqn:grad_RBM_b}\\
\frac{\partial l_s(\theta)}{\partial c_j}&=\frac{1}{N}\sum_{\mu = 1}^N \psi_s\big(\lambda_j(\ve{v}^{(\mu)},\theta)\big) -\ave{h_j}_s,
\label{eqn:grad_RBM_c}\\
\frac{\partial l_s(\theta)}{\partial w_{i,j}}&=\frac{1}{N}\sum_{\mu = 1}^N \mrm{v}_i^{(\mu)} \psi_s\big(\lambda_j(\ve{v}^{(\mu)},\theta)\big) -\ave{v_ih_j}_s,
\label{eqn:grad_RBM_w}
\end{align}
where $\ave{\cdots}_s$ is the expectation of RBM, i.e., 
\begin{align*}
\ave{A(\bm{v},\bm{h})}_s:=\sum_{\bm{v} \in \{-1,+1\}^{|V|}} \sum_{\bm{h} \in \mcal{X}(s)^{|H|}} A(\bm{v},\bm{h}) P_s(\bm{v}, \bm{h} \mid \theta),
\end{align*}
and 
\begin{align}
\psi_s(x):=\frac{\partial}{\partial x}\ln \phi_s(x)=
\begin{dcases}
\frac{s + 1}{s \tanh \{(s + 1)x/s\}} - \frac{1}{s \tanh(x/s)} & 1 \leq  s < \infty \\
\frac{1}{\tanh x} - \frac{1}{x} & s \to \infty
\end{dcases}
.
\label{eqn:psi_s(x)}
\end{align}
The log-likelihood function can be maximized by a gradient ascent method with the gradients expressed in Eqs. (\ref{eqn:grad_RBM_b})--(\ref{eqn:grad_RBM_w}). 
However, the evaluation of the expectations, $\ave{\cdots}_s$, included in the above gradients is computationally hard. 
The computation time of the evaluation grows exponentially as the number of variables increases.
Therefore, in practice, an approximate approach is used, for example, CD~\cite{CD2002}, pseudo-likelihood~\cite{RBM-PLE2010}, 
composite likelihood~\cite{RBM-CLE2012}, Kullback-Leibler importance estimation procedure (KLIEP)~\cite{RBM-KLIEP2011}, 
and Thouless-Anderson-Palmer (TAP) approximation~\cite{RBM-TAP2015}.
In particular, the CD method is the most popular method. 
In the CD method, the intractable expectations in Eqs. (\ref{eqn:grad_RBM_b})--(\ref{eqn:grad_RBM_w}) are approximated by 
the sample averages of the sampled points in which each sampled point is generated from the (one-time) Gibbs sampling 
using Eqs. (\ref{eqn:CondDistribution_V|H}) and (\ref{eqn:CondDistribution_H|V}), starting from each data point $\ve{v}^{(\mu)}$.

\subsection{Numerical experiment using artificial data}
\label{sec:RBM_experiment}

In the numerical experiments in this section, we used two RBMs: the generative RBM (gRBM), $P_1^{\mrm{gen}}$, and the learning RBM (tRBM), $P_s^{\mrm{train}}$. 
We obtained $N = 200$ artificial training data points, $D_V$, from the gRBM using Gibbs sampling and subsequently, we trained the tRBM using the data points. 
The sizes of the visible layers of both RBMs were the same, namely, $|V| = 8$.
The sizes of the hidden layers of the gRBM and tRBM were set to $|H| = 4$ and $|H| = 4 + R$, respectively. 
The sample space of the hidden variables in the gRBM was $\mcal{X}(1) = \{-1,+1\}$, implying that the gRBM is the binary-RBM.
The parameters of gRBM were randomly drawn: $b_i,c_j \sim G(0,0.1^2)$ and 
\begin{align}
w_{i,j}\sim U [-\sqrt{6/(|V|+|H|)},\sqrt{6/(|V|+|H|)}]
\label{eqn:XavierInitialization}
\end{align}
(Xavier's initialization~\cite{Xavier2010}), 
where $G(\mu,\sigma^2)$ is the Gaussian distribution and $U[\mrm{min},\mrm{max}]$ is the uniform distribution.

We trained the tRBM using the CD method. In the training, the parameters of tRBM were initialized by $b_i = c_j = 0$ and Eq. (\ref{eqn:XavierInitialization}). 
In the gradient ascent method, we used the full batch learning with the Adam method~\cite{Adam2015}. 
The quality of learning was measured using the Kullback-Leibler divergence (KLD) between the gRBM and tRBM: 
\begin{align}
\mrm{KLD}:=\frac{1}{|V|}\sum_{\bm{v} \in \{-1,+1\}^{|V|}}P_1^{\mrm{gen}}(\bm{v}) \ln \frac{P_1^{\mrm{gen}}(\bm{v})}{P_s^{\mrm{train}}(\bm{v})}.
\label{eqn:KLD}
\end{align}
The KLD is regarded as the (pseudo) distance between the gRBM and tRBM. 
Thus, it is a type of generalization error. 
We can evaluate the KLD (the generalization error) 
and log-likelihood function in Eq. (\ref{eqn:LogLikelihood}) (the negative training error) 
because the sizes of the RBMs are not large.

\begin{figure}[tb]
\centering
\includegraphics[height=4cm]{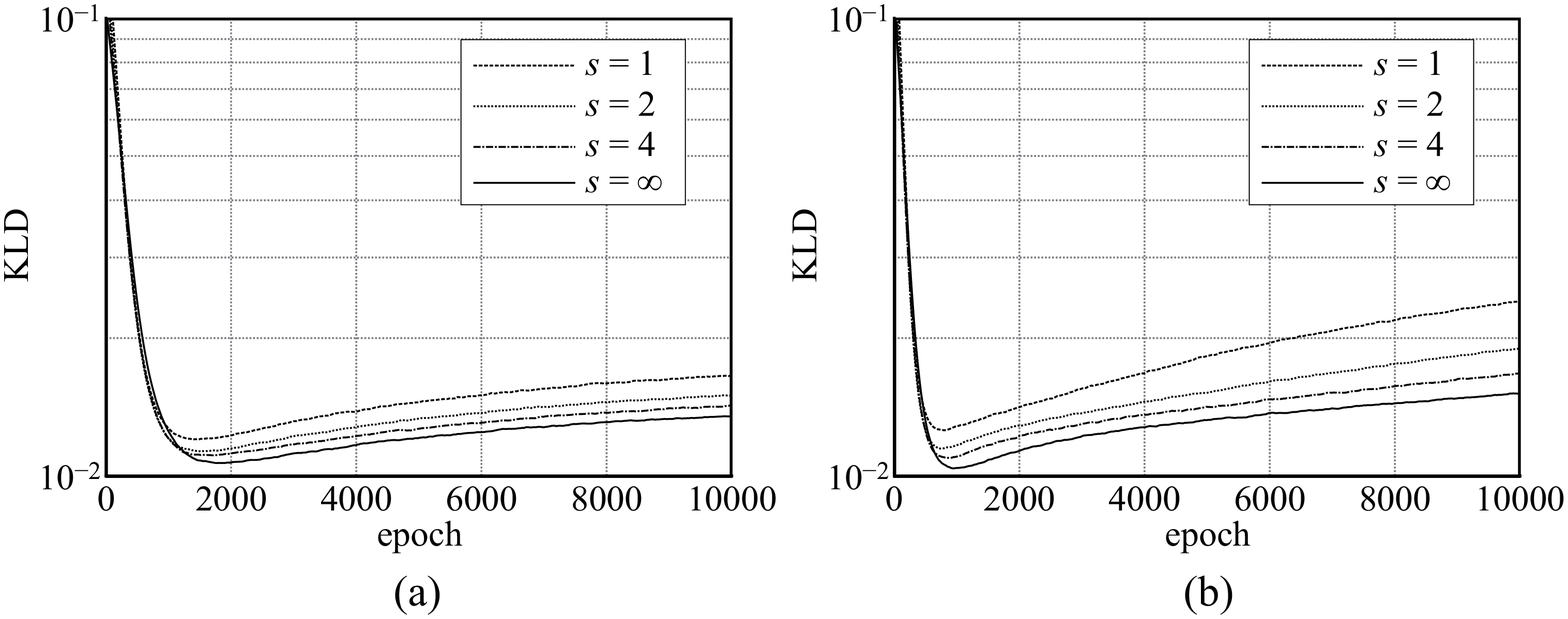}
\caption{KLDs against the number of parameter updates (epochs) when (a) $R = 0$ and (b) $R=5$. 
We used the tRBM with $s = 1,2,4,\infty$. These plots show the average over 300 experiments.}
\label{fig:KLD_CD_N200}
\end{figure}
\begin{figure}[tb]
\centering
\includegraphics[height=4cm]{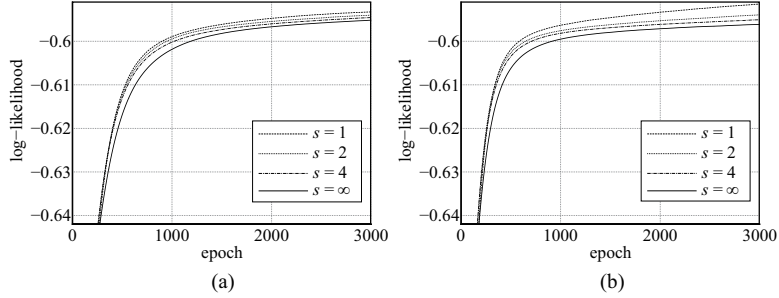}
\caption{Log-likelihoods (divided by $|V|$) against the number of parameter updates (epochs) when (a) $R = 0$ and (b) $R=5$. 
We used the tRBM with $s = 1,2,4,\infty$. These plots show the average over 300 experiments.}
\label{fig:LL_CD_N200}
\end{figure}
Figures \ref{fig:KLD_CD_N200} (a) and (b) show the KLDs against the number of parameter updates (i.e., the number of gradient ascent updates). 
We observe that all KLDs increase as the learnings proceed owing to the effect of over-fitting.
In Fig. \ref{fig:KLD_CD_N200} (a), because the gRBM and tRBM have the same structure (in other words, there is no model error), 
the effect of over-fitting is not severe. 
In contrast, in Fig. \ref{fig:KLD_CD_N200} (b), 
because the tRBM is more flexible than the gRBM, the effect of over-fitting tends to become severe. 
In fact, in Fig. \ref{fig:KLD_CD_N200} (b), the KLDs increase more rapidly as the learnings proceed. 
The increase in the KLD of higher $s$ is evidently slower. 
Figures \ref{fig:LL_CD_N200} (a) and (b) show the log-likelihood functions divided by $|V|$ against the number of parameter updates.
We observe that the log-likelihood function with lower $s$ grows more rapidly. 
In other words, the training error in the tRBM with lower $s$ decreases more rapidly. 
These results indicate that the multivalued hidden variables suppress over-fitting. 
In these experiments, the tRBM with $s = \infty$ is the best in terms of generalization.

\subsection{Effect of multivalued hidden variables}
\label{sec:ToyExample}

In the numerical experiments described in the previous section, we demonstrated that the multivalued hidden variables suppress over-fitting. 
In this section, we provide an insight into the effect of multivalued hidden variables using a toy example. 
Although the consideration presented below is for a simple RBM, which is significantly different from practical RBMs, 
it is expected to provide an insight into the effect of multivalued hidden variables.

First, let us consider a simple RBM with two visible variables:
\begin{align}
P_s(v_1,v_2, \bm{h} \mid w) = \frac{\omega(s)}{Z_s(w)}\exp\Big(w \sum_{i=1}^2\sum_{j \in H} v_ih_j\Big).
\label{eqn:Toy_RBM}
\end{align}
The marginal distribution of Eq. (\ref{eqn:Toy_RBM}) is 
\begin{align}
P_s(v_1,v_2 \mid w)=\frac{\exp\big\{ |H| \ln \phi_s \big(w(v_1 + v_2)\big)\big\}}
{ 2\exp\big\{ |H| \ln \phi_s(2w)\big\} + 2 \exp\big( |H| \ln 2\big)},
\label{eqn:Toy_marginal_RBM}
\end{align}
where we used $\lim_{x \to 0}\phi_s(x) = 2$ and $\phi_s(x) = \phi_s(-x)$. 
Because $v_1, v_2 \in \{-1,+1\}$, Eq. (\ref{eqn:Toy_marginal_RBM}) can be expanded as
\begin{align}
P_s(v_1,v_2 \mid w) =\frac{1+ m_{s,1}(w) v_1 + m_{s,2}(w) v_2 + \alpha_s(w)v_1v_2}{4},
\label{eqn:Toy_marginal_RBM_expand}
\end{align}
where 
\begin{align*}
m_{s,i}(w) &:= \sum_{v_1,v_2 \in \{-1,+1\}} v_i P_s(v_1,v_2 \mid w)  = 0, \\
\alpha_s(w)&:= \sum_{v_1,v_2 \in \{-1,+1\}} v_1v_2 P_s(v_1,v_2 \mid w)\\
&\>=\frac{\exp\big\{ |H| \ln \phi_s(2w)\big\} -  \exp\big( |H| \ln 2\big)}
{\exp\big\{ |H| \ln \phi_s(2w)\big\} +  \exp\big( |H| \ln 2\big)}\geq 0.
\end{align*}
Next, we consider the empirical distribution of $D_V$: 
\begin{align*}
Q_{D_V}(v_1,v_2):= \frac{1}{N}\sum_{\mu=1}^N \prod_{i = 1}^2\delta\big(v_i, \mrm{v}_i^{(\mu)}\big),
\end{align*}
where $\delta(x,y)$ is the Kronecker delta function. 
Similar to Eq. (\ref{eqn:Toy_marginal_RBM_expand}), the empirical distribution is expanded as
\begin{align}
Q_{D_V}(v_1,v_2) =\frac{1+ d_1 v_1 + d_2 v_2 + \beta v_1v_2}{4},
\label{eqn:empiricalDist_expand}
\end{align}
where $d_i := \sum_{\mu=1}^N \mrm{v}_i^{(\mu)} / N$ and $\beta := \sum_{\mu=1}^N \mrm{v}_1^{(\mu)}\mrm{v}_2^{(\mu)} / N$. 
For simplicity, in the following discussion, we assume that $d_1 = d_2 = 0$ and $\beta \geq 0$. 
Under this assumption, using the expanded forms in Eqs. (\ref{eqn:Toy_marginal_RBM_expand}) and (\ref{eqn:empiricalDist_expand}), 
the log-likelihood function of the simple RBM is expressed by
\begin{align}
l_s(w) &= \sum_{v_1,v_2 \in \{-1,+1\}}Q_{D_V}(v_1,v_2)\ln P_s(v_1,v_2 \mid w) \nn
&=\sum_{v_1,v_2 \in \{-1,+1\}}\frac{1 + \beta v_1 v_2}{4}\ln \frac{1 + \alpha_s(w)v_1v_2}{4}.
\label{eqn:LogLikelihood_ToyRBM}
\end{align}

\begin{figure}[tb]
\centering
\includegraphics[height=4cm]{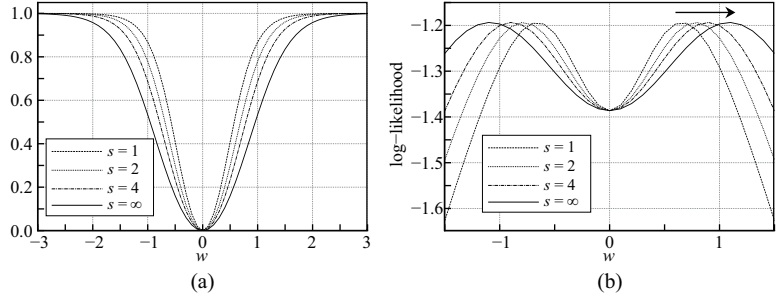}
\caption{(a) Plot of $\alpha_s(w)$ versus $w$ for various $s$ when $|H| = 2$. 
For $\beta = 0.6$, the values of $|w_s^*|$ for $s = 1,2,4$, and $\infty$ are approximately 0.6585, 0.7834, 0.8941, and 1.0887, respectively. 
(b) Plot of the log-likelihood function versus $w$ for various $s$ when $|H| = 2$ and $\beta = 0.6$. 
The shape of the peak around the global maximum point becomes sharper as $s$ decreases.}
\label{fig:result_toyRBM_H2}
\end{figure}

Ultimately, the aim of the maximum likelihood estimation is to find the value of $w$ that realizes $P_s(v_1,v_2 \mid w) = Q_{D_V}(v_1,v_2)$   
or in other words, to find a value of $w_s^*$ that satisfies $\alpha_s(w_s^*) = \beta$. 
The log-likelihood function in Eq. (\ref{eqn:LogLikelihood_ToyRBM}) is globally maximized at $w = w_s^*$ 
and the RBM with $w_s^*$ over-fits the data distribution. 
It can be shown that the function $\alpha_s(w)$ has the following three properties: 
(i) it is symmetric with respect to $w$,   
(ii) it monotonically increases with an increase in $w \geq 0$,   
and (iii) it monotonically decreases with an increase in $s$ when $|x| \not=0$. 
The function $\alpha_s(w)$ with $|H| = 2$ is shown in Fig. \ref{fig:result_toyRBM_H2} (a) as an example.
These three properties lead to the inequality $|w_s^*| < |w_{s+1}^*|$ for a certain $\beta > 0$, 
which implies that the global maximum point of the log-likelihood function in Eq. (\ref{eqn:LogLikelihood_ToyRBM}) moves away from the origin, $w = 0$, as $s$ increases (see Fig. \ref{fig:result_toyRBM_H2} (b)).

Usually, the initial value of $w$ is set to a value around the origin. 
As shown in Fig. \ref{fig:result_toyRBM_H2} (b), the global maximum point moves closer to the origin  
and the peak becomes sharper (in other words, the global maximum point becomes the stronger attractor) as $s$ decreases. 
This implies that with a gradient ascent type of algorithm, 
the RBM with a lower $s$ can reach the global maximum point more rapidly 
and causes over-fitting during an early stage of the learning. 
Whereas, the convergence with the global maximum point of the RBM with a higher $s$ is slower 
and it prevents over-fitting during an early stage of the learning
\footnote{Because the value of the log-likelihood function at the global maximum point for a higher $s$ is the same as that for a lower $s$, 
the RBM with a higher $s$ also causes over-fitting at that point.}. 
In fact, the increases in the generalization error (the KLD) and negative training error (the log-likelihood function) become faster as $s$ decreases 
in the numerical results obtained in the previous section (cf. Figs. \ref{fig:KLD_CD_N200} and \ref{fig:LL_CD_N200}).

From the above analysis, we found that the global maximum point moves away from the origin 
and becomes a weaker attractor as $s$ increases. 
This could lead to some expectations, for example:   
(i) in a more practical RBM, its log-likelihood function usually has several local maximum points,  
and thus, the RBM with a higher $s$ is more easily trapped by one of the local maximum points before converging with the global maximum point (namely, the over-fitting point) 
and (ii) some regularization methods, such as early stopping or $L_2$ regularization, are more effective in the RBM with a higher $s$.

\section{Application to classification problem}
\label{sec:PatternRecognitionApplication}

Let us consider a classification (or pattern recognition) problem in which an $n$-dimensional input vector $\bm{x} = (x_1, x_2,\ldots, x_n)^{\mrm{T}} \in \mathbb{R}^n$ is classified into $K$ different classes, 
$C_1, C_2, \ldots, C_K$. 
It is convenient to use a 1-of-$K$ representation (or a 1-of-$K$ coding) to identify each class~\cite{Bishop2006}.
In the 1-of-$K$ representation, each class corresponds to the $K$-dimensional vector $\bm{t} = (t_1, t_2,\ldots, t_K)^{\mrm{T}}$, 
where $t_k \in \{0,1\}$ and $\sum_{k = 1}^K t_k = 1$, i.e., $\bm{t}$ is a vector in which the value of only one element is one and the remaining elements are zero. 
When $t_k = 1$, $\bm{t}$ indicates class $C_k$. 
For simplicity of the notation, we denote the 1-of-$K$ vector, whose $k$th element is one, by $\bm{1}_k$.
In the following section, we consider the application of the proposed RBM to the classification problem.

\subsection{Discriminative restricted Boltzmann machine}
\label{sec:DRBM}

\begin{figure}[tb]
\centering
\includegraphics[height=2.5cm]{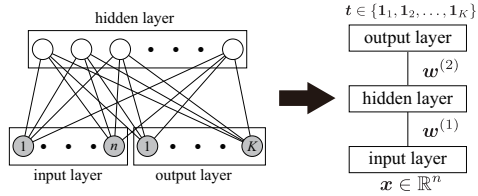}
\caption{Discriminative restricted Boltzmann machine is obtained to an extension of the RBM. 
Because the output layer corresponds to the 1-of-$K$ vector, it takes only $K$ different states. 
For distinction, the couplings between the input and hidden layers are represented by $\bm{w}^{(1)}$ and those between the hidden and output layers are represented by $\bm{w}^{(2)}$.}
\label{fig:DRBM}
\end{figure}

A discriminative restricted Boltzmann machine (DRBM) was proposed to solve the classification problem~\cite{DRBM2008,DRBM2012},  
which is a conditional distribution of the output 1-of-$K$ vector $\bm{t}$ conditioned with a continuous input vector $\bm{x}$.
The conventional DRBM can be obtained by a simple extension to the binary-RBM. 
The DRBM is obtained by the following process.  
The visible variables in the RBM are divided into two layers, the input and output layers.
The $K$ visible variables assigned to the output layer, $\bm{t}$, are redefined as the 1-of-$K$ vector with $\bm{1}_k$ as its realization 
(i.e., $\bm{t} \in \{\bm{1}_k \mid k = 1,2,\ldots, K\}$) 
and the $n$ visible variables assigned to the input layer, $\bm{x}$, are redefined as the continuous input vector (see Fig. \ref{fig:DRBM}).
Subsequently, we make a conditional distribution conditioned with the variables in the input layer: $P(\bm{t}, \bm{h} \mid \bm{x})$. 
Finally, by marginalizing the hidden variables out, we obtain the DRBM: $P(\bm{t}\mid \bm{x}) = \sum_{\bm{h}}P(\bm{t}, \bm{h} \mid \bm{x})$. 

By using the proposed RBM instead of the binary-RBM, 
we obtain an extension to the conventional DRBM, i.e., we obtain a DRBM with multivalued hidden variables.   
The proposed DRBM for $s \in \mathbb{N}$ is obtained by
\begin{align}
P_s(\bm{t} \mid \bm{x},\theta)&:=\frac{1}{\mcal{Z}_s(\bm{x},\theta)}
\exp\Big(\sum_{k=1}^K b_kt_k + \sum_{j \in H} \ln \phi_s\big(\zeta_j(\bm{t},\bm{x}, \theta) \big)\Big).
\label{eqn:DRBM}
\end{align}
where $\zeta_j(\bm{t},\bm{x}, \theta):= c_j + \sum_{k=1}^K w_{j,k}^{(2)}t_k + \sum_{i=1}^n w_{i,j}^{(1)}x_i$ and 
\begin{align}
\mcal{Z}_s(\bm{x},\theta):=\sum_{k=1}^K
\exp\Big(b_k + \sum_{j \in H} \ln \phi_s\big(\zeta_j(\bm{1}_k,\bm{x}, \theta) \big)\Big).
\label{eqn:PartitionFunction_DRBM}
\end{align}
The function $\phi_s(x)$ appearing in Eqs. (\ref{eqn:DRBM}) and (\ref{eqn:PartitionFunction_DRBM}) is already defined in Eq. (\ref{eqn:phi_s(x)}).
It is noteworthy that when $s = 1$, Eq. (\ref{eqn:DRBM}) is equivalent to the conventional DRBM proposed in Ref. \cite{DRBM2008}. 
Eq. (\ref{eqn:DRBM}) is regarded as the class probability, indicating that 
$P_s(\bm{t} = \bm{1}_k \mid \bm{x},\theta)$ is the probability of the input $\bm{x}$ belonging to class $C_k$. 
The input $\bm{x}$ should be assigned into a class that gives the maximum class probability.

Given $N$ supervised training data points, $D:=\{(\ve{t}^{(\mu)} ,\ve{x}^{(\mu)}) \mid \mu = 1,2,\ldots, N\}$,
the log-likelihood function of the proposed DRBM in Eq. (\ref{eqn:DRBM}) is defined as
\begin{align}
l_s^{\dagger}(\theta):=\frac{1}{N}\sum_{\mu=1}^N \ln P_s(\ve{t}^{(\mu)} \mid \ve{x}^{(\mu)},\theta).
\label{eqn:LogLikelihood_DRBM}
\end{align}
The gradients of the log-likelihood function with respect to the parameters are obtained as follows. 
\begin{align}
\frac{\partial l_s^{\dagger}(\theta)}{\partial b_k}&=\frac{1}{N}\sum_{\mu = 1}^N \big[ \mrm{t}_k^{(\mu)} - P_s(\bm{1}_k \mid \ve{x}^{(\mu)},\theta)\big],
\label{eqn:grad_DRBM_b}\\
\frac{\partial l_s^{\dagger}(\theta)}{\partial c_j}&=\frac{1}{N}\sum_{\mu = 1}^N \big[\psi_s\big(\zeta_j(\ve{t}^{(\mu)},\ve{x}^{(\mu)}, \theta)\big) 
-\ave{\psi_s\big(\zeta_j(\bm{t},\ve{x}^{(\mu)}, \theta)}_{\bm{t}}^{(\mu, s)} \big],
\label{eqn:grad_DRBM_c}\\
\frac{\partial l_s^{\dagger}(\theta)}{\partial w_{i,j}^{(1)}}&=\frac{1}{N}\sum_{\mu = 1}^N \mrm{x}_i^{(\mu)} \big[\psi_s\big(\zeta_j(\ve{t}^{(\mu)},\ve{x}^{(\mu)}, \theta)\big) 
-\ave{\psi_s\big(\zeta_j(\bm{t},\ve{x}^{(\mu)}, \theta)}_{\bm{t}}^{(\mu, s)}\big],
\label{eqn:grad_DRBM_w1} \\
\frac{\partial l_s^{\dagger}(\theta)}{\partial w_{j,k}^{(2)}}&=\frac{1}{N}\sum_{\mu = 1}^N \psi_s\big(\zeta_j(\bm{1}_k,\ve{x}^{(\mu)}, \theta)\big)
\big[ \mrm{t}_k^{(\mu)} - P_s(\bm{1}_k \mid \ve{x}^{(\mu)},\theta)\big],
\label{eqn:grad_DRBM_w2}
\end{align}
where $\ave{\cdots}_{\bm{t}}^{(\mu, s)}$ denotes the expectation defined by
\begin{align*}
\ave{A(\bm{t})}_{\bm{t}}^{(\mu, s)}:=\sum_{k=1}^K A(\bm{1}_k) P_s(\bm{1}_k \mid \ve{x}^{(\mu)},\theta).
\end{align*}
The function $\psi_s(x)$ appearing in the above gradients is already defined in Eq. (\ref{eqn:psi_s(x)}). 
It is noteworthy that the gradients expressed in Eqs. (\ref{eqn:grad_DRBM_b})--(\ref{eqn:grad_DRBM_w2}) are computed without an approximation, unlike those in the RBM,  
owing to the special structure of DRBM. 
In the training, we maximize $l_s^{\dagger}(\theta)$ with respect to $\theta$ using a gradient ascent method with Eqs. (\ref{eqn:grad_DRBM_b})--(\ref{eqn:grad_DRBM_w2}).

\subsection{Numerical experiment using MNIST data set}
\label{sec:DRBM_experiment}

\begin{figure}[tb]
\centering
\includegraphics[height=4cm]{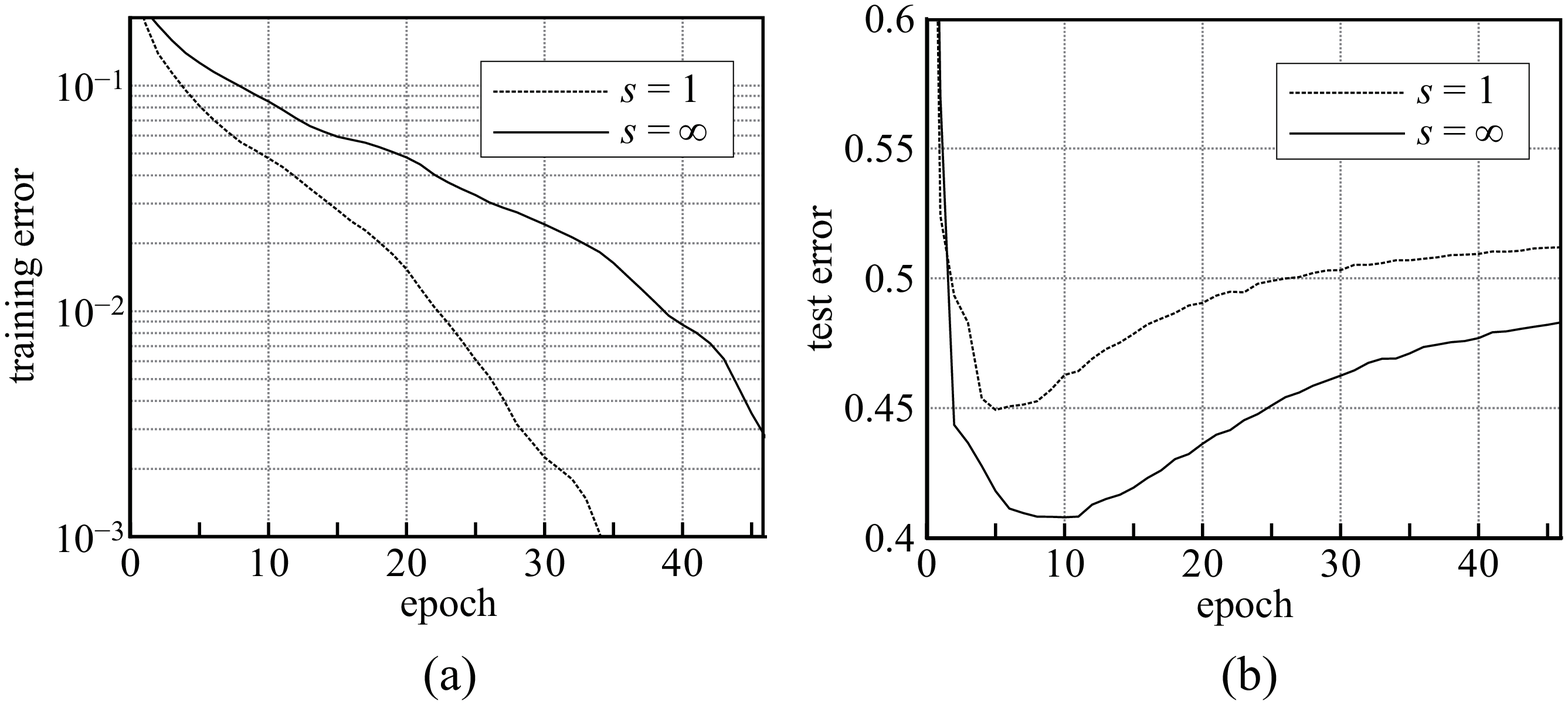}
\caption{Missclassification errors against the number of parameter updates (epochs): (a) training error and (b) test error. 
Here, one epoch consists of one full update cycle over the training data set, implying that one epoch involves $N /B = 10$ updates by the SGA in this case. 
We used the DRBM with $s = 1,\infty$. These plots show the average over 120 experiments.}
\label{fig:DRBM_MNIST}
\end{figure}
In this section, we show the results of the numerical experiment using MNIST.
MNIST is a data set of 10 different handwritten digits, $0, 1, \ldots,$ and $9$, 
and is composed of $60000$ training data points and $10000$ test data points. 
Each data point includes the input data, a $28 \times 28$ digit (8-bit) image, and the corresponding target digit label. 
Therefore, for the data set, we set $n = 784$ and $K = 10$. 
All input images were normalized by dividing by $255$ during preprocessing. 

We trained the proposed DRBM with $|H|=200$ using $N = 1000$ training data points in MNIST and tested it using 10000 test data points. 
In the training, we used the stochastic gradient ascent (SGA), for which the mini-batch size was $B = 100$, with the AdaMax optimizer~\cite{Adam2015}.
All coupling parameters were initialized by the Xavier method~\cite{Xavier2010} 
and all bias parameters were initialized by zero. 
Figure \ref{fig:DRBM_MNIST} shows the plots of the missclassification rates for (a) training data set and (b) test data set  
versus the number of parameter updates. 
All input images in the test data set were corrupted by the Gaussian noise with $\sigma = 120$ before the normalization
\footnote{Each corrupted input image $\hat{\ve{x}}$ was created from the corresponding original image $\ve{x}$ by 
$\hat{\mrm{x}}_i = \mrm{x}_i + \epsilon_i$, where $\epsilon_i$ is the additive white Gaussian noise drawn from $G(0,120^2)$. 
If $\hat{\mrm{x}}_i > 255$, we set $\hat{\mrm{x}}_i = 255$ 
and if $\hat{\mrm{x}}_i < 0$, we set $\hat{\mrm{x}}_i = 0$.}. 
We observe that the DRBM with $s = \infty$ is better in terms of generalization 
because it shows a higher training error and lower test error.  
This indicates that the multivalued hidden variables are also effective in the DRBM.

\section{Conclusion}
\label{sec:conclusion}

In this paper, we proposed an RBM with multivalued hidden variables, which is a simple extension to the conventional binary-RBM, 
and showed that the proposed RBM is better than the binary-RBM in terms of the generalization property 
via numerical experiments conducted on CD learning with artificial data (in Sec. \ref{sec:RBM_experiment}) 
and classification problem with MNIST (in Sec. \ref{sec:DRBM_experiment}). 

It is important to understand the reason why the multivalued hidden variables are effective in terms of over-fitting. 
We provided a basic insight into it by analyzing a simple example in Sec. \ref{sec:ToyExample}. 
However, practical RBMs are much more complex than the simple example used in this study. 
Therefore, we need to perform further analysis to clarify this reason. 
We think that a mean-field analysis~\cite{NishimoriBook} can be used to perform the further analysis. 
Moreover, a criteria for over-fitting was provided in Ref.~\cite{Coolen2017}. 
The relationship between the criteria and our multivalued hidden variables is also interesting. 
These issues will be addressed in our future studies.

\subsubsection*{acknowledgment}
This work was partially supported by JSPS KAKENHI (Grant Numbers 15K00330, 15H03699, 18K11459, and 18H03303), 
JST CREST (Grant Number JPMJCR1402), and the COI Program from the JST (Grant Number JPMJCE1312).

\bibliographystyle{jpsj}
\bibliography{cite}

\end{document}